# Tournament selection in zeroth-level classifier systems based on average reward reinforcement learning

Zang Zhaoxiang, Li Zhao, Wang Junying, Dan Zhiping
zxzang@gmail.com; zangzx@hust.edu.cn
(Hubei Key Laboratory of Intelligent Vision Based Monitoring for Hydroelectric Engineering,
China Three Gorges University, Yichang Hubei, 443002, China;
College of Computer and Information Technology, China Three Gorges University, Yichang Hubei,
443002, China)

**Abstract:** As a genetics-based machine learning technique, zeroth-level classifier system (ZCS) is based on a discounted reward reinforcement learning algorithm, bucket-brigade algorithm, which optimizes the discounted total reward received by an agent but is not suitable for all multi-step problems, especially large-size ones. There are some undiscounted reinforcement learning methods available, such as R-learning, which optimize the average reward per time step. In this paper, R-learning is used as the reinforcement learning employed by ZCS, to replace its discounted reward reinforcement learning approach, and tournament selection is used to replace roulette wheel selection in ZCS. The modification results in classifier systems that can support long action chains, and thus is able to solve large multi-step problems.

# 1 Introduction

Learning Classifier Systems (LCSs) are rule-based adaptive systems which use Genetic Algorithm (GA) and some machine learning methods to facilitate rule discovery and rule learning[1]. LCSs are competitive with other techniques on classification tasks, data mining[2, 3] or robot control applications[4, 5]. In general, an LCS is a model of an intelligent agent interacting with an environment. Its ability to choose the best policy acting in the environment, namely adaptability, improves with experience. The source of the improvement is the learning from reinforcement, i.e. payoff, provided by the environment. The aim of an LCS is to maximize the achieved environmental payoffs. To do this, LCSs try to evolve and develop a population of compact and maximally general "condition-action-payoff" rules, called classifiers, which tell the system in each state (identified by the condition) the amount of payoffs for any available action. So, LCSs can be seen as a special method of reinforcement learning that provides a different approach to get generalization.

The original Learning Classifier System framework proposed by Holland, is referred to as the traditional framework now. And then, Willson proposed strength-based Zeroth-level Classifier System (ZCS)[6], and accuracy-based X Classifier System (XCS)[7]. The XCS classifier system has solved the former main shortcoming of LCSs, which is the problem of strong over-generals, by its accuracy based fitness approach. Bull and Hurst[8] have recently shown that, despite its relative simplicity, ZCS is able to perform optimally through its use of fitness sharing. That is, ZCS was shown to perform as well, with appropriate parameters, as the more complex XCS on a number of tasks.

Despite current research has focused on the use of accuracy in rule predictions as the fitness measure, the present work departs from this popular approach and takes a step backward, aiming to uncover the potential of strength based LCS (and particularly ZCS) in sequential decision problems. In this direction, we will discuss the use of average reward in ZCS, and will introduce an undiscounted reinforcement learning technique called R-learning[9, 10] for ZCS to optimize average reward, which is a different metric from the discounted reward optimized by original ZCS. In particular, we apply R-learning based ZCS to large multi-step problems and compare it with ZCS. Experimental results are encouraging, in that ZCS with R-learning can perform optimally or near optimally in these problems. Later, we will refer to our proposal as "ZCSAR", and the "AR" stands for "average reward".

The rest of the paper is structured as follows: Section 2 provides some necessary background knowledge on reinforcement learning, including Sarsa and R-learning. Section 3 provides a brief description of ZCS and maze environments. How ZCS can be modified to include the average reward reinforcement learning is described in Section 4, while Section 5 analyzes the trouble resulting from our modification to ZCS, and presents some solution to it. Experiments with our proposal and some related discussion are given in Section 6. Finally, Section 9 ends the paper, presents our main conclusions and some directions for future research.

## 2 Reinforcement learning

Reinforcement learning is a formal framework in which an agent manipulates its environment through a series of actions, and receives some rewards as feedback to its actions, but is not told what the correct actions would have been. The agent stores its knowledge about how to make decisions that maximize rewards or minimize costs over a period of time. Reinforcement learning must learn to perform a task by trial and error from a reinforcement signal (the reward values) that is not as informative as might be desired. In reinforcement learning for multi-step problems, the reinforcement signal usually gives delayed reward, which typically comes at the end of a series of actions. Delayed reward makes learning much more difficult.

Generally, the reinforcement learning framework consists of
- A discrete set of environment states, $S$;
- A discrete set of available actions, $A$;
- An immediate reinforcement function $R$, mapping $S \times A$ into the real value $r$, where $r$ is the expected environmental payoff after performing the action $a$, from $A$, in a particular state $s$, from $S$.

On each step of interaction the agent perceives the environment to be in state $s$; the agent then chooses an action $a$ in the set $A$, and the action $a$ is performed in the environment. As a result of taking action $a$, the agent receives a reward $r$ and a new state $s'$.

The agent's job is to find a policy $\pi$, mapping states to actions, that maximizes some long-run measure of reinforcement. There are mainly two measures to value a policy: discounted reward optimality and average reward optimality.

In discounted reinforcement learning, the performance measure being optimized usually is the infinite-horizon discounted model[11], which takes the long-run reward of the agent into account, but rewards receiving in the future are geometrically discounted according to a discount factor $\gamma (0 \leq \gamma < 1)$:

$$\lim_{N \to \infty} E\left(\sum_{t=0}^{N-1} \gamma^t r_t(s)\right) \tag{1}$$

where $E$ denotes expected value, and $r_t(s)$ is the reward received at time $t$ starting from state $s$ under a policy. An optimal discounted policy maximizes the above infinite-horizon discounted reward.

On the other hand, undiscounted reinforcement learning usually optimizes the average reward model[9], in which the agent is supposed to take actions that maximize its long-run average reward per step:

$$\rho(s) = \lim_{N \to \infty} \frac{E\left(\sum_{t=0}^{N-1} r_t(s)\right)}{N} \tag{2}$$

If a policy maximizes the average reward over all states, it is referred to as a gain optimal policy. Usually, average reward $\rho(s)$ can be denoted as $\rho$, which is state independent[12] and greatly simplifies the design of average reward algorithms.

How does the agent find a policy to maximize the long-run measure of reinforcement? Most of the reinforcement learning algorithms are based on estimating state-action pair value function (called action value function) that indicates how good it is for the agent to perform a given action in a given state. Here, "how good" is defined in terms of future expected reward value, usually as (1) or (2), corresponding to the discounted reward and average reward optimality. We will give a brief description of two typical reinforcement learning algorithms based on discounted reward and on average reward optimality, respectively.

## 2.1 Sarsa Algorithm

Sarsa is a well-known reinforcement learning algorithm that can be seen as a variant of Q-learning algorithm[11]. It is based on iteratively approximating the table of all action values $Q(s,a)$, named the Q-table. Initially, all the $Q(s,a)$ values are set to 0. At time step $t-1$, the agent perceives the environment state $s_{t-1}$, chooses an action $a_{t-1}$ by the $\varepsilon$-greedy policy. The action $a_{t-1}$ is performed in the environment, and the agent receives an immediate reward $r_{imm}(s_{t-1}, a_{t-1})$ for doing action $a_{t-1}$, and a new environment state $s_t$. Then, the entry $Q(s_{t-1}, a_{t-1})$ is updated using the following rule:

$$Q(s_{t-1}, a_{t-1}) \leftarrow Q(s_{t-1}, a_{t-1}) + \beta \left(\hat{Q}(s_{t-1}, a_{t-1}) - Q(s_{t-1}, a_{t-1})\right) \tag{3}$$

Here, $0 \leq \beta \leq 1$ is the learning rate controlling how quickly errors in the estimated action values are corrected; $\hat{Q}(s_{t-1}, a_{t-1})$ is the new estimate of $Q(s_{t-1}, a_{t-1})$, and is computed as

$$\hat{Q}(s_{t-1}, a_{t-1}) = r_{imm}(s_{t-1}, a_{t-1}) + \gamma Q(s_t, a), \tag{4}$$

where $r_{imm}(s_{t-1}, a_{t-1})$ is the immediate reward received for performing $a_{t-1}$ in state $s_{t-1}$.

## 2.2 R-learning

Since Q-learning discounts future rewards, it prefers actions that result in short-term ordinary rewards to those that result in long-term sustained or considerable rewards. On the contrary, the R-learning algorithm[9] proposed by Schwartz maximizes the average reward per time step.

R-learning is similar to Q-learning in form. It is based on iteratively approximating the action values $R(s,a)$, which represent the average adjusted reward of doing an action $a$ in state $s$ once, and then following corresponding policy subsequently.

R-learning algorithm consists of the following steps:

1) Initialize all the $R(s,a)$ values to zero, and the average reward variable $\rho$ also initialized to zero.

2) Let the current time step be $t-1$. From the current state $s_{t-1}$, choose an action $a_{t-1}$ by some exploration/action-selection mechanism, such as the $\varepsilon$-greedy policy.

3) Perform the action $a_{t-1}$, observe the immediate reward $r_{imm}(s_{t-1}, a_{t-1})$ received and the subsequent state $s_t$.

4) Update R values using the following rule:

$$R(s_{t-1}, a_{t-1}) \leftarrow R(s_{t-1}, a_{t-1}) + \beta_R \left( r_{imm}(s_{t-1}, a_{t-1}) - \rho + \max_{a \in A} R(s_t, a) - R(s_{t-1}, a_{t-1}) \right) \tag{5}$$

5) If $R(s_{t-1}, a_{t-1}) = \max_{a \in A} R(s_{t-1}, a)$ (i.e. if a greedy/non-random action $a_{t-1}$ was chosen), then update the average reward $\rho$ according to the rule:

$$\rho \leftarrow \rho + \beta_\rho \left( r_{imm}(s_{t-1}, a_{t-1}) + \max_{a \in A} R(s_t, a) - \max_{a \in A} R(s_{t-1}, a) - \rho \right) \tag{6}$$

6) $t \leftarrow t+1$, and go to step 2.

Here, $0 \leq \beta_R \leq 1$ is the learning rate for updating action values $R(\cdot, \cdot)$, and $0 \leq \beta_\rho \leq 1$ is the learning rate for updating average reward $\rho$.

The update rule for action value $R(\cdot, \cdot)$ differs from the rule for Q-learning in subtracting the average reward $\rho$ from the immediate reward, and not discounting the next maximum action value. The estimation of the average reward $\rho$ is a critical task in R-learning. As mentioned above, the average reward $\rho$, under some conditions, does not depend on any state, and is constant over the whole state space[12]. This facilitates the use of average reward algorithms.

Following the basic R-learning algorithm, [10] proposed some variations. The variations mainly focus on different ways to update the average reward, corresponding to the step 5 given above.

## 3 ZCS Classifier System and Its Testing Environments

### 3.1 A Brief Description of ZCS

The following is a brief description of ZCS, further information can be found in [6] and [8].

The ZCS architecture was introduced by Stewart Wilson in 1994. It is a Michigan style LCS without internal memory, which periodically receives a binary encoded input from its environment. The system determines an appropriate response based on this input and performs the indicated action, usually altering the state of the environment. The action is rewarded by a scalar reinforcement. Internally the system cycles through a sequence of performance, reinforcement and discovery.

The ZCS rule base consists of a population of classifiers, symbolized by $[P]$. This population has a fixed maximum size $N$. Each classifier is a condition-action-strength rule $<c,a,str>$. The rule condition $c$ is a string of characters from the ternary alphabet $\{0,1,\#\}$, where # acts as a wildcard allowing a classifier to generalize over different input messages. The action $a \in \{a_1,\cdots,a_n\}$ is represented by a binary string and both conditions and actions are initialized randomly. Strength scalar $str$ acts as an indication of the perceived utility of that rule within the system. The strength of each rule is initialized to a predetermined value termed $S_0$.

On receipt of an environmental input message $s_t$, the rule-base is scanned and any classifiers whose condition matches input message $s_t$ is placed in a match set [M]. Match set [M] is a subset of the whole population $[P]$ of classifiers. If on some time-step, [M] is empty or has a total strength $Str_{[M]}$ that is less than a fixed fraction $\phi(0<\phi\leq 1)$ of the mean strength of the population $[P]$, then a covering operator is invoked. A new rule is created with a condition that matches the environmental input and a randomly selected action. The rule's condition is then made less specific by the random inclusion of #'s at a probability of $P_\#$ per bit. The new rule is given a strength equal to the population average and inserted into the population, overwriting a rule selected for deletion. The deleted rules are chosen using roulette-wheel selection based on the reciprocal of strength.

Thus a particular action $a$ is selected from the match set by roulette wheel selection policy based on the total strength $Str(s_t,a)$ of the classifiers in [M] which advocate that action. For all actions $a \in \{a_1,\cdots,a_n\}$ in [M], $Str(s_t,a)$ is named as system strength, which is computed as:

$$Str(s_t,a) = \sum_{cl.a=a \wedge cl \in [M]} cl.str \qquad (7)$$

$cl$ stands for a classifier, $cl.str$ for strength of $cl$, and $cl.a$ for its action.

When an action has been selected, all rules in the [M] that advocate this action are placed in action set [A] and the system executes the action. Depending on environmental circumstances, a scalar reward reinforcement value $r$ (maybe null) is supplied to ZCS as a consequence of executing $a$, together with a new input configuration $s_{t+1}$.

Reinforcement in ZCS consists of redistributing payoff between subsequent action sets. In each cycle, a "bucket-brigade" credit-assignment policy similar to Sarsa is employed:

1) A fixed fraction $\beta (0 < \beta \leq 1)$ of the strength of each member of [A] at current time step is deducted and placed in a common bucket $B$: $str_{[A]}(i) = (1-\beta) \cdot str_{[A]}(i)$; $B = \beta \cdot \sum_i str_{[A]}(i)$, where $str_{[A]}(i)$ stands for the strength of the i-th classifier of [A]. $B$ is initially set to zero.

2) If a reward $r_{-1}$ is received from the environment as a consequence of executing $a_{-1}$ at the previous time step t-1, then a fixed fraction $\beta (0 < \beta \leq 1)$ of $r_{-1}$ is distributed evenly amongst the members of $[A]_{-1}$: $str_{[A]_{-1}}(i) = str_{[A]_{-1}}(i) + \beta r_{-1} / |A_{-1}|$, where $|A_{-1}|$ is the number of classifiers in $[A]_{-1}$.

3) Classifiers in $[A]_{-1}$ (if it is non-empty) have their strengths incremented by $\gamma B / |A_{-1}|$, $str_{[A]_{-1}}(i) = str_{[A]_{-1}}(i) + \gamma B / |A_{-1}|$, where $\gamma$ is a pre-determined discount factor ($0 < \gamma \leq 1$), $B$ is the total amount put in the current bucket in step 1.

4) Finally, the bucket $B$ is emptied, and all classifiers in the set difference [M] - [A] have their strengths reduced by a small fraction $\tau (0 < \tau < 1)$, which acts as a "tax" to encourage exploitation of strong classifier sets: $\forall cl \in [M] \land cl \notin [A]$ : $cl.str = (1-\tau)cl.str$.

Then the above process can be written as a re-assignment:

$$Str_{[A]_{-1}} \leftarrow Str_{[A]_{-1}} + \beta(r_{-1} + \gamma \cdot Str_{[A]} - Str_{[A]_{-1}}) \qquad (8)$$

$Str_{[A]_{-1}}$ is the total strength of members of $[A]_{-1}$, also known as $Str(s_{t-1}, a_{-1})$; $Str_{[A]}$ is the total strength of members of [A], also known as $Str(s_t, a)$. So, Equation can be rewritten as

$$Str(s_{t-1}, a_{-1}) \leftarrow Str(s_{t-1}, a_{-1}) + \beta(r_{-1} + \gamma \cdot Str(s_t, a) - Str(s_{t-1}, a_{-1})) \qquad (9)$$

ZCS employs GA as discovery mechanism over the whole rule-set [P] at each instance (panmictic). On each cycle there is a probability $\rho_{GA}$ of GA invocation. When called, the GA uses roulette wheel selection to determine the parent rules based on strength. Two offspring are produced via crossover (single point, using probability $\chi$) and mutation (using probability $\mu$). The parents then donate half their strength to their offspring who replace existing members of the population. The deleted rules are chosen based on the reciprocal of strength.

## 3.2 Maze Environments

Maze problems, usually represented as grid-like two-dimensional areas that may contain different objects of any quantity and with different properties (for example, obstacle, goal, or can be empty), serve as a simplified virtual model of the real environment, and can be used for developing core algorithms of many real-world applications related to the problem of navigation. The agent should learn the shortest path to goal states, without knowing the environmental model in advance.

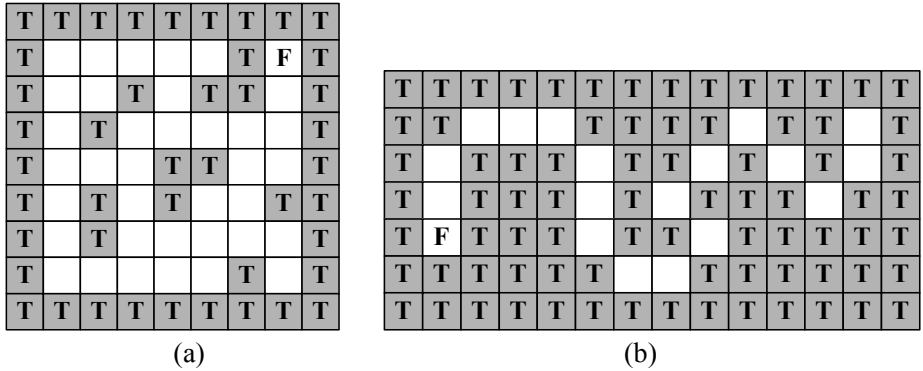

**Figure 1.** (a) Maze6 environment; (b) Woods14 environment. Food object is marked with F, and obstacle is marked with T.

LCS has been the most widely used class of algorithms for reinforcement learning in mazes for the last twenty years, and has presented the most promising performance results[6, 13]. Figure 3(a) presents Woods1[6] maze environment. The maze may contain different obstacles in any quantity, such as T standing for tree in Woods1, and some objects for learning purposes, like virtual food F, which is the agents' goal to reach. It must be noted that, if a maze has not enough obstacles to mark its boundary, the left and right edges of the maze are connected, as are the top and bottom. In this paper, the agent is randomly placed in the maze on an empty cell, and the agent has two boolean sensors for each of the eight adjacent squares. The agent can move into any adjacent square that is free.

## 4 Adding R-learning to ZCS

In this section, we show how ZCS can be modified to include R-learning[9, 10] to optimize average reward, which is different from the discounted reward optimized by Sarsa-learning. The implementation of our system, ZCSAR, is also discussed here.

As mentioned above, ZCS uses a "bucket-brigade" credit-assignment policy similar to Sarsa to update the classifiers population. From Equation (9), bucket-brigade algorithm in ZCS is indeed similar to the Sarsa update rule (3). Besides, the comparison shows that (i) ZCS represents each entry in the Q-table by a set of classifiers, i.e. $Q(s_{t-1}, a_{t-1})$ is represented by the classifiers in $[A]_{-1}$, and $Q(s_t, a_t)$ is represented by the classifiers in $[A]$; (ii) The system strength $Str(s_t, a)$, specified in Equation (7), also known as $Str_{[A]}$, corresponds to the value $Q(s_t,a)$ in Equation (3), and $r_{-1}+\gamma \cdot Str(s_t,a)$ in Equation (9) corresponds to the estimate $\hat{Q}(s_{t-1},a_{t-1})$ of value $Q(s_{t-1},a_{t-1})$ in Equation (4); (iii) Only one entry $Q(s_{t-1},a_{t-1})$ is updated in tabular Sarsa algorithm at time step $t$, while in ZCS a set of classifiers is usually updated in one time step.

R-learning has been introduced in Section 2, and it is a new type of reinforcement learning. R-learning and Sarsa algorithm are similar in form but not in meaning, since Sarsa algorithm is based on the discounted reward optimality, while R-learning, based on the average reward optimality, maximizes the average reward per step. In R-learning, we can define the estimate of $R(s_{t-1},a_{t-1})$ as

$$\hat{R}(s_{t-1},a_{t-1}) = r_{imm}(s_{t-1},a_{t-1}) - \rho + \max_{a \in A} R(s_t,a) \qquad (10)$$

Thus, Equation (5) can be rewritten as

$$R(s_{t-1},a_{t-1}) \leftarrow R(s_{t-1},a_{t-1}) + \beta_R \left(\hat{R}(s_{t-1},a_{t-1}) - R(s_{t-1},a_{t-1})\right) \qquad (11)$$

The major difference between Equation (11) and (3) is that they use different methods to compute the estimate $\hat{R}(s_{t-1},a_{t-1})$ and $\hat{Q}(s_{t-1},a_{t-1})$. Additionally, R-learning needs to estimate the average reward $\rho$, which is extra work than in Sarsa algorithm.

From what has been discussed above, the analogies between Sarsa and ZCS, the difference and similarity between Sarsa and R-learning have been presented. We can get that, the system strength $Str(s_t,a)$ in ZCS corresponds to the action value $R(s_t,a)$, and $r_{-1}+\gamma \cdot Str(s_t,a)$ in ZCS corresponds to the new estimate $\hat{R}(s_{t-1},a_{t-1})$ of $R(s_{t-1},a_{t-1})$. In order to add R-learning to ZCS, we only need to focus on the methods to compute $\hat{R}(s_{t-1},a_{t-1})$ in Equation (10) and $r_{-1}+\gamma \cdot Str(s_t,a)$ in Equation (9). Given the correspondence between the system strength $Str(s_t,a)$ and the action value $R(s_t,a)$, the average reward approach to compute $r_{-1}+\gamma \cdot Str(s_t,a)$ in Equation (9) can be modified as $r_{-1} - \rho + Str(s_t,a)$. Thus, Equation (9) is changed as:

$$Str(s_{t-1},a_{-1}) \leftarrow Str(s_{t-1},a_{-1}) + \beta(r_{-1} - \rho + Str(s_t,a) - Str(s_{t-1},a_{-1})) \qquad (12)$$

Equation (12) will replace Equation (9) in ZCS to change the whole reinforcement learning mechanism employed by the original ZCS.

About the specific update rule of classifiers in $[A]_{-1}$, the step 2 and 3 in Section 3.1 can be modified as:

2) If a reward $r_{-1}$ is received from the environment as a consequence of executing $a_{-1}$ at the previous time step t-1, and the estimate of average reward is $\rho$, then a fixed fraction $\beta(0 < \beta \leq 1)$ of $r_{-1} - \rho$ is distributed evenly amongst the members of $[A]_{-1}$: $str_{[A]_{-1}}(i) = str_{[A]_{-1}}(i) + \beta(r_{-1} - \rho)/|A_{-1}|$, where $|A_{-1}|$ is the number of classifiers in $[A]_{-1}$.

3) Classifiers in $[A]_{-1}$ (if it is non-empty) have their strengths incremented by $B/|A_{-1}|$, $str_{[A]_{-1}}(i) = str_{[A]_{-1}}(i) + B/|A_{-1}|$, where $B$ is the total amount put in the current bucket.

Next, a procedure to estimate the average reward $\rho$ needs to be added to ZCS. Step 5 in the description of R-learning algorithm in Section 2.2 can be moved to ZCS through some modifications. To do so, Step 5 in Section 2.2 can be rewritten as:

If $Str_{[A]_{-1}} = \max_{a \in A} Str(s_{t-1}, a)$ (i.e. if a greedy/non-random action $a_{t-1}$ was chosen), then update the average reward $\rho$ according to the rule:

$$\rho \leftarrow \rho + \beta_\rho \left( r_{-1} + \max_{a \in A} Str(s_t, a) - \max_{a \in A} Str(s_{t-1}, a) - \rho \right) \tag{13}$$

The new type of Step 5 can be inserted into the procedure of ZCS, and located just before the update of classifiers in $[A]_{-1}$. It must be noted that, at the first time step of each trial in an experiment, there is no need to update the average reward $\rho$, since no previous environmental reward available at that time. And at the beginning of an experiment, $\rho$ is initialized to zero. In addition, the update value of average reward $\rho$ is not used in Equation (12) directly. Instead, its more stable moving average value is adopted to avoid the heavy oscillations with its update values, since average reward $\rho$ is updated by the immediate reward $r_{t-1}$ which is stochastic and with great fluctuation. The window size for moving average is 100, i.e. moving average is computed as the average of the last 100 updated values. If the window size is too small, the moving average will have no effect; if the window size is too big, the changing trend of average reward will be hidden, which will limit the immediate feedback function of average reward.

Through the two steps above, we have replaced Sarsa algorithm in ZCS with R-learning, getting the new system ZCSAR. However, in order to speed up the process of convergence in ZCSAR, the fluctuation of the estimate $\rho$ needs to be reduced over time. So we make the learning rate $\beta_\rho$ in Equation (13) decayed over time using a simple rule:

$$\beta_\rho \leftarrow \beta_\rho - \frac{\beta_\rho^{\max} - \beta_\rho^{\min}}{NumOfTrials}, \tag{14}$$

where $\beta_\rho^{\max}$ is the initial value of $\beta_\rho$, $\beta_\rho^{\min}$ is the minimum learning rate required, and *NumOfTrials* is the number of exploration trials (problems) in an experiment. $\beta_\rho$ is updated at the beginning of each exploration trial using Equation (14), but not at each time step.

## 5 Subtraction Trouble and Tournament Selection

When ZCSAR uses Equation (12) as reinforcement learning mechanism, some issues arise. The update rule for system strength $Str(\cdot, \cdot)$ differs from the rule for Sarsa-learning in subtracting the average reward $\rho$ from the immediate reward, and not discounting the next system strength (action value). The subtraction may cause system strength $Str(\cdot, \cdot)$ negative, which does not appear in original ZCS and discounted reward reinforcement learning Sarsa. The negative $Str(\cdot, \cdot)$, less than zero, occurs when the value of $r_{-1} - \rho + Str(s_t, a)$ is continuously negative for some time steps. In most time steps, reward is delayed, so $r_{-1}$ is zero. The estimation of the average reward $\rho$ is not an easy task in sparse reward domains. It may differ largely from the true value of average reward in early stage of learning. Thus, whether the value of $r_{-1} - \rho + Str(s_t, a)$ is negative or not depends mainly on the difference of $\rho$ and $Str(s_t, a)$.

If $Str(\cdot, \cdot)$ is negative, the sum of strength of classifiers in action set is also less than zero, which means some classifiers' strength is negative in action set. However, all components of ZCS were designed on the supposition that classifier's strength is greater than zero. Specially, roulette wheel selection (proportionate selection) based on classifier's strength (or its reciprocal) is adopted as action selection method in match set [M], parents selection method in GA, classifier selection method in GA deletion and covering operator deletion. It is known that classifier's strength must be positive in roulette wheel selection. ZCS is in line with this requirement, but not ZCSAR.

This is a problem caused by subtraction. To address this problem, an easy way is to make negative values be zero, i.e. let classifier's strength not less than zero. We indicate this method as "truncation". In other words, if ZCSAR still uses roulette wheel selection, truncation is an easy method to adapt it.

However, is truncation method proper and effective for ZCSAR? Is there any alternative to tackle this problem? A promising proposal is to replace roulette wheel selection with tournament selection in ZCSAR. Tournament selection with tournament sizes proportionate to the actual set size is shown to outperform roulette wheel selection in the widely-used classifier system XCS[14]. So it is expected that tournament selection can also improve the performance of ZCSAR. And importantly, in contrast to roulette wheel selection, tournament selection is independent of fitness scaling and does not require positive classifier strength, so classifiers' strength can be less than zero in ZCSAR with tournament selection.

In tournament selection, classifiers are not selected proportional to their strength, but tournaments are held in which the classifier with the highest strength wins. Stochastic tournaments are not considered herein. Participants for the tournament are chosen at random from the corresponding classifier set in which selection is applied. The size of the tournament is dependent on the corresponding classifier set size, and the size of each tournament has the size of the fraction $\theta \in (0,1]$ of the corresponding classifier set size. Parameter $\theta$ controls the selection pressure. Instead of roulette wheel selection in action selection in match set [M], parents selection in GA, classifier deletion selection in GA and covering operator, three independent tournaments are held in which the classifier with the highest (or lowest) strength is selected, and $\theta$ values are 0.1, 0.4, 0.6 respectively.

Later, we will refer to our proposals as "ZCSAR+Roulette" and "ZCSAR+Tournament" in the remainder of this work, to indicate ZCSAR with roulette wheel selection and truncation method, and ZCSAR with tournament selection respectively.

## 6 Experiments in Maze Environments

Two maze problems are tested and studied here, to illustrate the generality and effectiveness of our approaches, and ZCS for comparison.

## 6.1 Experimental Setup

To conduct experiments, every experiment typically consists of 12000 problems (trials) that the agent must solve. And for each problem, the agent is placed into a randomly chosen empty square in the mazes. Then the agent moves under the control of the classifier system avoiding obstacles until either it reaches the food or had taken 500 steps, at which point the problem ended unconditionally. The agent will not change its position if it chooses an action to move to a square with an obstacle inside, though one time-step still elapses. When the agent reaches the food, it receives a constant reward of 1000; otherwise, it receives a reward equal to 0. And in order to evaluate the final policy evolved, in each experiment, exploration is turned off during the last 2000 problems and the system works only in exploitation. In exploitation problems, the action which predicts the highest payoff is always selected in match set [M], and the genetic algorithm is turned off. System performance is computed as the average number of steps to food in the last 50 problems. Every statistic results presented in this paper is averaged on 10 experiments.

The following classifier structure was used for LCS in the experiments: Each classifier has 16 binary bits in the condition field: two bits for each of the 8 neighbouring squares, with 00 representing the situation that the square is empty, 11 that it contains food (F), and 01 that it is an obstacle (T).

the general LCS's parameters used for ZCS, ZCSAR+Roulette, and ZCSAR+Tournament are set as follows: β=0.6, $\tau$=0.1, $\rho_{GA}$=0.25, $\phi$=0.5, χ=0.5, μ=0.002, $S_0$=20.0, $P_\#$=0.33, $N$=800. Some specific parameters are set as follows: for ZCSAR+Roulette, and ZCSAR+Tournament, $\beta_\rho^{max}$=0.005, $\beta_\rho^{min}$=0.00001; and in ZCS, $\gamma$=0.71. The detailed description of these parameters is available in [6] and [8].

## 6.2 Experimental Results and Discussions

In the first experiment, we applied ZCS, ZCSAR+Roulette and ZCSAR+Tournament to Maze6 environment (Figure 1(a)). Maze6 is a typical and somewhat difficult environment for testing the learning systems since the goal position for agents to reach is hidden by some obstacles, and there is not any regularity in it. Each sensory-action pair in this maze almost needs a special classifier to cover (i.e. it only allows few generalizations), so ZCS is likely to produce over-general classifiers in it. Besides, the optimal solution in Maze6 requires the agent to perform long sequences of actions to reach the goal state. The optimal average path to the food in Maze6 is 5.19 steps. This experiment is used to show that ZCS with average reward reinforcement learning can solve the general maze problem.

Figure 2 reports the performance of ZCS, ZCSAR+Roulette and ZCSAR+Tournament in Maze6 environment. In the three cases, the results all converge to near optimum during the last 2000 exploitation problems, and there is almost no difference between them, about 5.85, 6.21, and 6.02 respectively. ZCSAR+Roulette and ZCSAR+Tournament can almost perform as well as ZCS in this environment. During the learning period (first 10000 problems), the three systems' performance deviates from optimum, since the GA continues to function and probabilistic action selection (roulette wheel selection or tournament selection) is used. In addition, ZCSAR+Tournament changes continuously and oscillates heavily within the first 10000 learning problems, which is possibly caused by tournament selection used as action selection mechanism in match set [M].

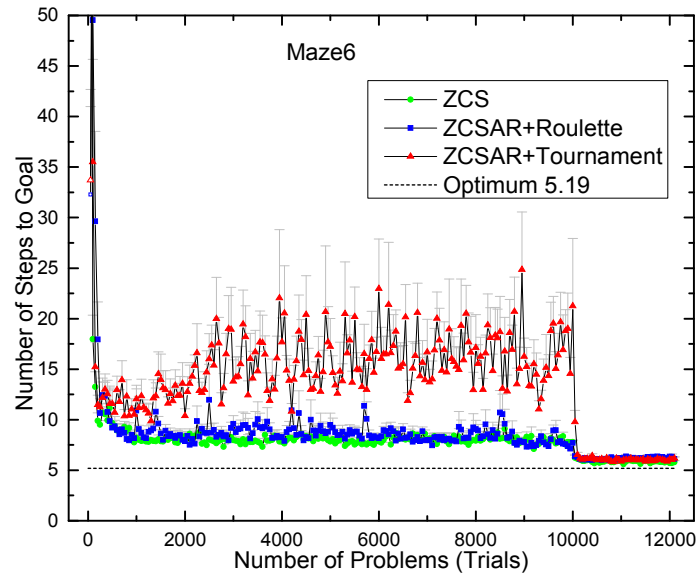

**Figure 2.** Performance of applying ZCSAR+Roulette and ZCSAR+Tournament to Maze6, compared with ZCS. Error bars represent the standard error. Curves are averages over 10 experiments.

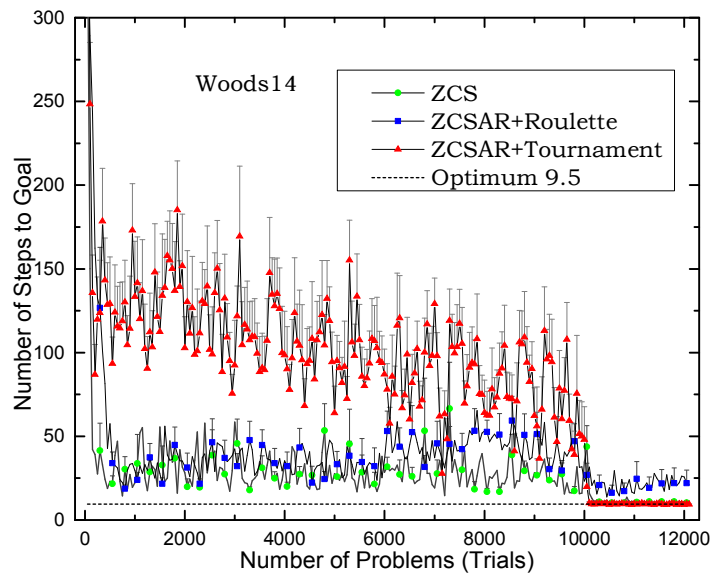

**Figure 3.** Performance of applying ZCSAR+Roulette and ZCSAR+Tournament to Woods14, compared with ZCS. Error bars represent the standard error. Curves are averages over 10 experiments.

In the second experiment, the testing environment is Woods14 (Figure 1(b)), which is a corridor of 18 blank cells and a food cell at the end. The optimal average path to the food in Woods14 is 9.5 steps. The agent needs longer sequences of actions to reach the goal position, resulting in a sparser reception of delayed reward. So, it is complex to most LCSs[15].

It can be seen from Figure 2 that, in Woods14, performances of the three systems oscillate above the optimum during training period, while evolve promising solutions during the last 2000 exploitation problems. ZCSAR+Tournament gets about 9.50 steps to find food, and ZCS gets about 10.70 steps. ZCSAR+Roulette performs less well (near optimum) and converges to about 21.36 steps. ZCSAR+Tournament can get the optimal solution in Woods14. This seems because of the average reward reinforcement learning and tournament selection employed by ZCSAR+Tournament, which guarantees the system can disambiguate those early states in the long action chains effectively.

## 7 Conclusions

In this paper, due to the similarity between Sarsa and bucket-brigade algorithm in ZCS, and the similarity in form between Sarsa algorithm and R-learning, bucket-brigade algorithm in ZCS is replaced with R-learning through some modifications. R-learning is an undiscounted reinforcement learning technique to optimize average reward, which is a different metric from the discounted reward optimized by bucket-brigade algorithm. Thus, ZCS with R-learning, ZCSAR, is able to maximize the average reward per time step, not the cumulative discounted rewards. This is helpful to support long action chains in large multi-step learning problems.

However, R-learning will cause some classifiers' strength is negative in ZCSAR. This does not meet the supposition that classifier's strength is greater than zero in ZCS. Specially, roulette wheel selection based on classifier's strength (or its reciprocal) used in ZCS requires that classifier's strength is positive. To address this problem, two extended systems are presented: "ZCSAR+Roulette" and "ZCSAR+Tournament". ZCSAR+Roulette indicates ZCSAR with roulette wheel selection and truncation method, while ZCSAR+Tournament indicates ZCSAR with tournament selection. Truncation means to cut off those negative strength values, set them to zero.

We test ZCSAR+Roulette and ZCSAR+Tournament on two well-known multi-step problems, compared with ZCS. Overall, experiments show that ZCSAR+Tournament can evolve optimal or near-optimal solutions in these typically difficult multi-step environments, while ZCSAR+Roulette can just reach the suboptimum in Woods14 environment. Especially in Woods14 environment, the performance of ZCSAR+Tournament is very good, but ZCS just reaches a near-optimal performance.

Because of the basic change of the reinforcement learning employed by ZCS, and tournament selection is used to replace roulette wheel selection, ZCSAR+Tournament still needs some extra testing to study their performance in other problems. Additionally, we plan to consider the impact of average reward reinforcement learning in ZCS when the environment is stochastic.

# References:

[1]. Bull, L., A brief history of learning classifier systems: from CS-1 to XCS and its variants. Evolutionary Intelligence, 2015: p. 1-16.
[2]. Ebadi, T., et al., Human-interpretable Feature Pattern Classification System using Learning

Classifier Systems. Evolutionary Computation, 2014. 22(4): p. 629-650.

[3]. Tzima, F.A. and P.A. Mitkas, ZCS Revisited: Zeroth-Level Classifier Systems for Data Mining, in Proceedings of the 2008 IEEE International Conference on Data Mining Workshops. 2008, IEEE Computer Society: Washington, DC, USA. p. 700--709.

[4]. Cadrik, T. and M. Mach, Control of agents in a multi-agent system using ZCS evolutionary classifier systems, in 2014 IEEE 12th International Symposium on Applied Machine Intelligence and Informatics (SAMI). 2014, IEEE: Herl'any, Slovakia. p. 163-166.

[5]. Cádrik, T. and M. Mach, Usage of ZCS Evolutionary Classifier System as a Rule Maker for Cleaning Robot Task, in Emergent Trends in Robotics and Intelligent Systems, P. Sinčák, et al., P. Sinčák, et al.^Editors. 2015, Springer International Publishing. p. 113-119.

[6]. Wilson, S.W., ZCS: A zeroth level classifier system. Evolutionary Computation, 1994. 2(1): p. 1-18.

[7]. Wilson, S.W., Classifier Fitness Based on Accuracy. Evolutionary Computation, 1995. 3(2): p. 149-175.

[8]. Bull, L. and J. Hurst, ZCS Redux. Evolutionary Computation, 2002. 10(2): p. 185-205.

[9]. Schwartz, A., A reinforcement learning method for maximizing undiscounted rewards, in Proceedings of the Tenth International Conference on Machine Learning, P. Utgoff, P. Utgoff^Editors. 1993, Morgan Kaufmann. p. 298-305.

[10]. Singh, S.P., Reinforcement learning algorithms for average-payoff Markovian decision processes, in Proceedings of the twelfth national conference on Artificial intelligence (vol. 1). 1994, American Association for Artificial Intelligence: Menlo Park, CA, USA. p. 700--705.

[11]. Sutton, R.S. and A.G. Barto, Reinforcement learning: an introduction. Adaptive computation and machine learning. 1998, Cambridge, MA: MIT Press.

[12]. Mahadevan, S., Average reward reinforcement learning: Foundations, algorithms, and empirical results. Machine Learning, 1996. 22: p. 159-195.

[13]. Zatuchna, Z. and A. Bagnall, A learning classifier system for mazes with aliasing clones. Natural Computing, 2009. 8(1): p. 57-99.

[14]. Butz, M.V., K. Sastry and D.E. Goldberg, Strong, Stable, and Reliable Fitness Pressure in XCS due to Tournament Selection. Genetic Programming and Evolvable Machines, 2005. 6(1): p. 53-77.

[15]. Zang, Z., et al., Learning classifier system with average reward reinforcement learning. Knowledge-Based Systems, 2013. 40(0): p. 58 - 71.